%% file: naaclhlt2019.tex
\documentclass[11pt,a4paper]{article}
\usepackage{acl2019}
\usepackage{times}
\usepackage{latexsym}
\usepackage{amsmath}
\usepackage{tikz,booktabs,float,color,adjustbox}

\usepackage{url}
\aclfinalcopy 
\def\aclpaperid{77} 

\title{ Improving Zero-shot  Translation with Language-Independent Constraints}

\author{Ngoc-Quan Pham$^*$ \and Jan Niehues$^+$ \and Thanh-Le Ha$^+$ \and Alex Waibel$^+$ \\
		$^*$ Karlsruhe Institute of Technology \\
        {\tt \{ngoc.pham, thanh-le.ha, alex.waibel\}@kit.edu} \\
		$^+$ Maastricht University \\
        \tt jan.niehues@maastrichtuniversity.nl}

\date{}

\begin{document}

\maketitle
\begin{abstract}

An important concern in training multilingual neural machine translation (NMT) is to translate between language pairs unseen during training, i.e zero-shot translation. Improving this ability kills two birds with one stone by providing an alternative to pivot translation which also allows us to better understand how the model captures information between languages.

In this work, we carried out an investigation on this capability of the multilingual NMT models. First, we intentionally create an encoder architecture which is independent with respect to the source language. Such experiments shed light on the ability of NMT encoders to learn multilingual representations, in general. Based on such proof of concept, we were able to design regularization methods into the standard Transformer model, so that the whole architecture becomes more robust in zero-shot conditions. We investigated the behaviour of such models on the standard IWSLT 2017 multilingual dataset. We achieved an average improvement of $2.23$ BLEU points across $12$ language pairs compared to the zero-shot performance of a state-of-the-art multilingual system. Additionally, we carry out further experiments in which the effect is confirmed even for language pairs with multiple intermediate pivots. 




\end{abstract}

\section{Introduction}
Neural machine translation (NMT) exploits neural networks to directly learn to transform sentences from a source language to a target language~\cite{Sutskever2014,Bahdanau2014}. Universal multilingual NMT discovered that a neural translation system can be trained on datasets containing source and target sentences in multiple languages~\cite{firat2016multi,Johnson2016}. Successfully trained models using this approach can be used to translate arbitrarily between any languages included in the training data. In low-resource scenarios, multilingual NMT has proven to be an extremely useful regularization method since each language direction benefits from the information of the others~\cite{Ha2016,gu2018universal}. 

An important research focus of multilingual NMT is zero-shot translation (ZS), or translation between languages included in multilingual data for which no directly parallel training data exists. Application-wise, ZS offers a faster and more direct path between languages compared to pivot translation, which requires translation to one or many intermediate languages. This can result in large latency and error propagation, common issues in non-end-to-end pipelines.
From a representation learning point of view, there is evidence of NMT's ability to capture language-independent features, which have proved useful for cross-lingual transfer learning~\cite{zoph2016transfer,kim2019transfer} and provide motivation for ZS translation. However it is still unclear if minimizing the difference in representations between languages is beneficial for zero-shot learning.

On the other hand, the current neural architecture and learning mechanisms of multilingual NMT is not geared towards having a common representation. Different languages are likely to convey the same semantic content with sentences of different lengths~\cite{kalchbrenner2016neural}, which makes the desiderata difficult to achieve. Moreover, the loss function of the neural translation model does not favour having sentences encoded in the same representation space regardless of the source language. As a result, if the network capacity is large enough, it may partition itself into different sub-spaces for different language pairs~\cite{arivazhagan2019missing}.    

Our work here focuses on the zero-shot translation aspect of universal multilingual NMT. First, we attempt to investigate the relationship of encoder representation and ZS performance. By modifying the Transformer architecture of \citet{vaswani2017attention} to afford a fixed-size representation for the encoder output, we found that we can significantly improve zero-shot performance at the cost of a lower performance on the supervised language pairs. To the best of our knowledge, this is the first empirical evidence showing that the multilingual model can capture both language-independent and language-dependent features, and that the former can be prioritized during training. 

This observation leads us to the most important contribution in this work, which is to propose several techniques to learn a joint semantic space for different languages in multilingual models without any architectural modification. The key idea is to prefer a source language-independent representation in the decoder using an additional loss function. As a result, the NMT architecture remains untouched and the technique is scalable to the number of languages in the training data. The success of this method is shown by significant gains on zero-shot translation quality in the standard IWSLT 2017 multilingual benchmark~\cite{cettolo2017overview}. 
Finally, we introduce a more challenging scenario that involves more than one bridge language between source and target languages. This challenging setup confirms the consistency of our zero-shot techniques while clarifying the disadvantages of pivot-based translation.

\section{Background: Multilingual Neural Machine Translation}
Given an input sequence $X$ and its translation $Y$, neural machine translation (NMT) uses sequence-to-sequence models~\cite{Sutskever2014} to directly model the posterior probability of generating $Y$ from $X$. 

Universal multilingual NMT expands the original bilingual setting by combining  parallel corpora from multiple language pairs into one single corpus. By directly training the NMT model on this combined corpus, the model can be made to translate sentences from any seen source language into any seen target language. Notably, this multilingual framework does not yield any difference in the training objective, i.e maximizing the likelihood of the target sentence $Y$ given the source sentence $X$:

\begin{equation}
Loss(X, Y) = - P (Y|X)
\label{eq:nmt}
\end{equation}

Previous work on universal NMT proposed different methods to control language generation. While source language identity may not be the concern, the decoder requires a target language signal to generate sentences in any desired language. Work from~\citet{Ha2016} and~\citet{Johnson2016} used the addition of language identity tokens in order to minimize architectural changes while controlling generation.
Subsequently, stronger constraints were bestowed upon the decoder to force the correct language to be generated through language features or vocabulary filtering during decoding~\cite{ha2017effective}. 

In practice, the number of language pairs in a multilingual corpus increases exponentially over the size of the language set. Therefore, a multilingual corpus rarely covers all of the language pairs involved, resulting in a need to investigate translation between the missing directions. The missing directions are referred as `zero-shot translation' as the model has no access to any explicit parallel samples, naturally or artificially. 


\section{Proof of concept: Fixed-size encoder representations for language-independence}
\label{sec:pooling}

\begin{figure*}[ht!]
\vspace{-1em}
\centering
\includegraphics[width = 0.9\textwidth]{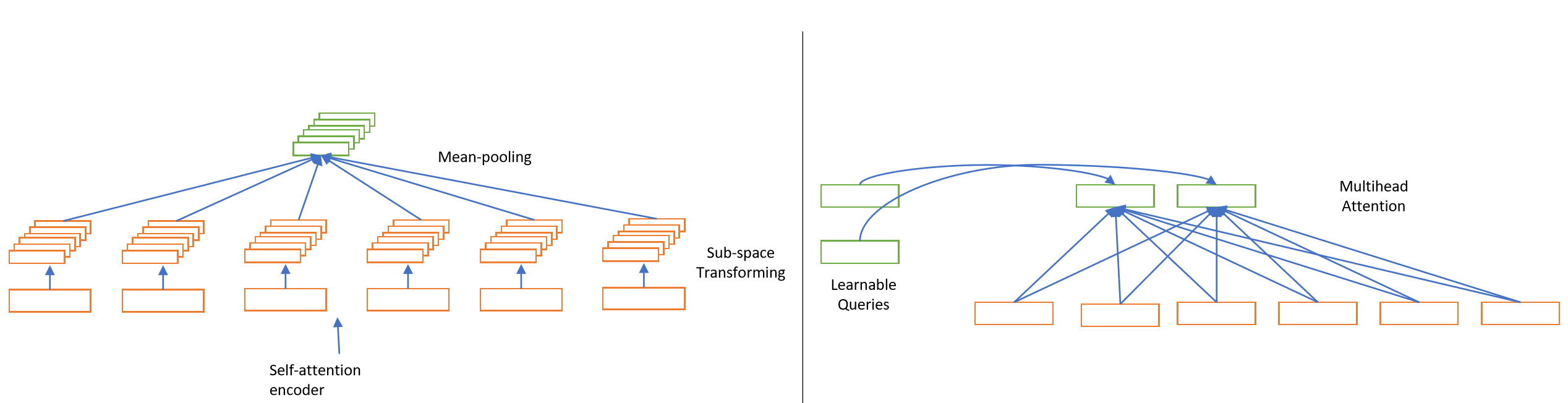}
\caption{\label{fig:pooling} Fixed-size representations using multi-head mean-pooling (left) and attention-pooling (right).}
\end{figure*}

As the length of encoder representations depends on the source language, current architectures are not ideal to learn language-independent encoder representations. Therefore, we propose different architectures with fixed-size encoder representations. This also allows us to directly compare encoder representations of different languages, and to enforce such similarity through an additional loss function. This modification comes with the price of an information bottleneck due to the process of removing the length variability. On the other hand, it adds additional regularization which would naturally prioritize the features shared between languages. 

Motivated by the literature in sentence embeddings~\cite{schwenk2017learning,wang-etal-2017-sentence}, we take the average over time of the encoder states. 
Specifically, assume that $X$ is the set of source embeddings input to the encoder: 
\begin{equation}
\begin{array}{lr}
H_v = Encoder(X) \\
H_f = mean\_pooling(H_v)
\end{array}
\label{eq:sent_emb}
\end{equation}
    
The purpose of this modification is two-fold. First, this model explicitly opens more possibilities for language-independent representation to occur, because every sentence is compressed into a consistent number of states. Second, we can observe the balance between language-independent and language-dependent information in the encoder; if zero-shot performance is minimally affected, then the encoder is in general able to capture language-independent information, and this restricted encoder retains this information. 

However, this model naturally has a disadvantage due to the introduced information bottleneck, similar to non-attention models~\cite{Sutskever2014,kalchbrenner2013recurrent}. We alleviate this problem by expanding the number of hidden states of the encoder output. As a result we investigate two variations of pooling as follows:  


\paragraph{Multi-head Mean-Pooling}
While taking the average over time significantly reduces the model capacity, we can allocate more capacity for the model by linearly projecting the variable-length representation. By concatenating the pooled values from different sub-spaces, we obtain a fixed-size representation with the size $N\times H$. 
However, instead of learning to pay attention to input tokens normally, this decoder learns to distribute its focus into each mean-pooled embedding. 

\paragraph{Multi-head Attention-Pooling}
The attention model is notable for its ability to extract relevant information from a sequence, which is an alternative to using pooling operators. However, self-attention is not within our architectural choices because the self-attention output has the same number of states with the input, while we need to restrict to a fixed set. We instead propose to set a fixed number of queries as learnable parameters for the model, so it will learn to extract necessary information from the sequence to include in the limited space. It is possible for this model to converge to mean-pooling because these parameters are not as informative as either encoder or decoder states. However, our experiments later on have proven this does not occur in practice. 

These two variations are illustrated in Figure~\ref{fig:pooling}. Here we investigate these models for the purpose of observing the relationship between encoder representations and zero-shot performance. Section \ref{experiments} shows that, despite the fact that this model falls short against the baseline Transformer in non-zero-shot tests, we observed that the retained information in the bottleneck does not affect the performance of zero-shot translation, our motivation for the upcoming objectives. 
\paragraph{Language-Independent Objective}
With constant length encoder output, we can now design an objective function using this advantage for language-independent representation. Hypothetically, for true multi-parallel data in which sentences from different languages are aligned, we can force the encoder outputs to be the same for the aligned sentences (newly enabled by the fixed state size). In other multilingual frameworks in which each data sample is bilingual, we exploit the fact that certain languages are shared between multiple language pairs (in order to enable zero-shot translation). As a result, by using such languages as a bridge, we can simply minimize the squared root deviation (i.e Min Squared Error - MSE) of the encoder representations between the bridge and other languages, and adding the regularization term $R(X, Y)$ to the loss function:
\begin{equation}
\begin{array}{lr}
Loss(X, Y) = - P (Y|X) + \alpha (R(X, Y)) \\
R(X, Y) = - (Encoder(X) - Encoder(Y))^2
\end{array}
\label{eq:enc_bridge}
\end{equation}

In Equation~\ref{eq:enc_bridge}, we ran the encoder on both languages. Assuming sentence $X$ belongs to the bridge language, the loss function will lead to the similar representation of different sentences in other languages that were aligned with $X$. The difficulty lies in optimizing two objectives at once: the second acts as a regularization because it prevents language-specific information from being included in the encoder output. As well, many multilingual corpora may not contain perfectly aligned sentences, which is a hindrance for language bridging. 

\section{Source Language-Independent Decoders}
\label{sec:att-bridge}
We have so far described our proposed method to learn language-independent features. We introduce the fixed-size states for the encoder and adds a regularization term to the NMT loss function to encourage similarity between encoder states. The problem with this method is the limiting factor of the fixed-size representations. With the standard architecture, while the length of the encoder states always depends on the source sentence, at each timestep the decoder only has access to a fixed representation of the encoder (context vector from attention). This observation suggests that forcing a decoder state to be independent of the source language and maintaining the variable-size representation for the encoder is possible. In this section, we navigate the target NMT architecture back to the popular variable-length sequential encoder in which no such compromise was made. 
    
Starting from the above motivation, the key idea is to force a source language-independent representation in the decoder using an additional loss function. We achieve this by operating the encoder-decoder flow not only from the source sentence to the target, but also from the source to itself to recreate the source sentence. While this resembles an auto-encoder which can be combined with translation~\cite{he2016dual,domhan2017using}, it is not necessary to minimize the auto-encoder likelihood as in the multi-task approach~\cite{niehues2017exploiting}, but only the decoder-level similarity between the true target sentence and the auto-encoded source sentence. Due to the lack of true parallel data, this method serves as a bridge between the different languages. 

An important feature of the NMT attention mechanism is that it extracts relevant information in encoded memory (the keys and queries, in this case they are the source sentence hidden states) and compresses them into one single state. More importantly, in the decoder operation this operator dynamically repeats every timestep. By using the encoder to encode both (source and target) sentences and operate the attentive decoder on top of both encoded sentences, we obtain two attentive representations of the two sentences which are equally long. This is the key to enabling forced-length representations in our model. 

\begin{figure*}[htb]
\vspace{-1em}
\centering
\includegraphics[width = 0.75\textwidth]{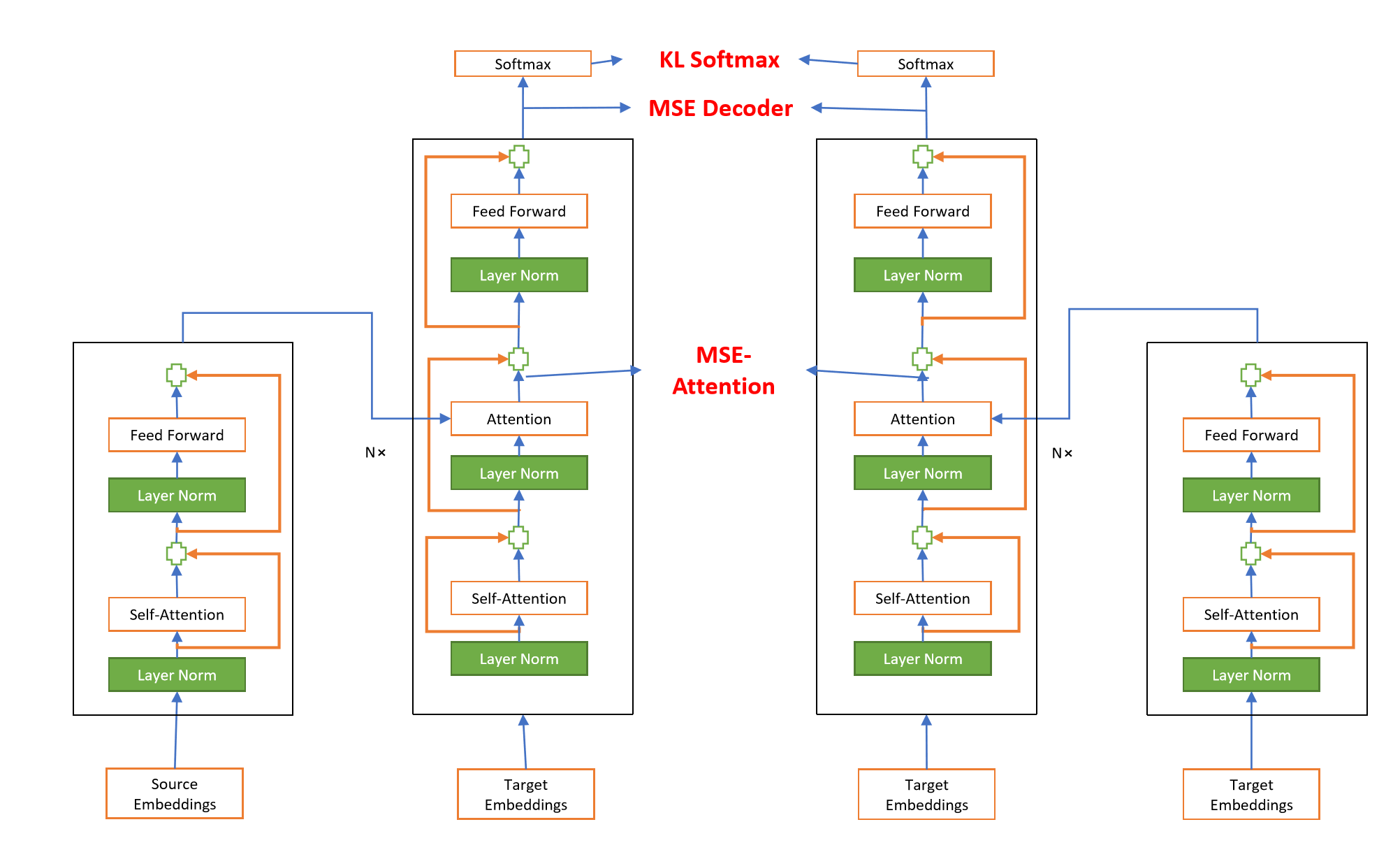}
\caption{\label{fig:method} Three different constraints for language-independent decoders. The model is run twice as translation (left) and auto-encoder (right). The KL-Softmax is applied at the very top, while the MSE-Decoder minimizes difference between the layer-normalized states at the end of the decoder. The MSE-attention operates on non-normalized attention outputs. }
\end{figure*}

Given the described model, the question is about where in the model we can apply our representation-forcing from Equation~\ref{eq:enc_bridge}. Due to the nature of many translation models being multi-layered, it is not as straightforward as in the pooled encoder models. Hence, we investigate three different locations where this regularization method can be applied. Their illustration is depicted in Figure~\ref{fig:method}.

\paragraph{Attention Forcing} We can force each attention context vector\footnote{As in~\cite{luong2015effective} the context vector denotes the weighted sum of the encoder after the attention operation} to be the same between two decoding outputs. As a Transformer has $N$ decoder layers, we take $N$ attention vectors of each decoder and apply MSE element-wise\footnote{For multi-head attention we take the output after concatenating the heads.}. This \textbf{MSE-Attention} method is naturally the most immediate derivative of forcing encoder states to be similar. Here we annotate $Att^n(Y_t, X)$ to be the context vector from attention of layer $n$ between decoder state of token $Y_t$ and source sentence $X$.
\begin{equation}
R(X, Y) = -\sum_{n=0}^{N-1} \sum_{t=0}^{T-1}(Att^n(Y_t, X) - Att^n(Y_t, Y))^2
\label{eq:mse-attn}
\end{equation}

\paragraph{Decoder Forcing} Instead of optimizing for each context vector, we can also regularize the final decoder layer (before the Softmax layer), because this state summarizes the information gathered in the decoder at each timestep. This approach will be referred as \textbf{MSE-Decoder}. Similar to Equation~\ref{eq:mse-attn}, we denote the decoder state of the conditional probability at time $t$ as $Dec(Y_t|X, Y_{1..t-1})$.

\begin{equation}
\begin{split}
R(X, Y) = \sum_{t=0}^{T-1}( & Dec(Y_t|X, Y_{1..t-1}) \\ 
& - Dec(Y_t|Y, Y_{1..t-1}))^2
\end{split}
\label{eq:mse-dev}
\end{equation}

\paragraph{Softmax Forcing} Similar to our second variation, the restriction is now put at the final layer. By running the decoder twice for translation and auto-encoder, we can force the output distribution of each step $P(Y_t|X, Y_{1..t-1})$ to be equal using KL divergence minimization. The purpose of this step is to enable the  decoder to generate the same target sentence with source sentences in different languages. We denote this approach as \textbf{KL-Softmax}.

\begin{equation}
\begin{split}
R(X, Y) =  \sum_{t=0}^{T-1} KL & (P(Y_t|X, Y_{1..t-1}), \\ 
&  P(Y_t|Y, Y_{1..t-1}))
\end{split}
\label{eq:klsoftmax}
\end{equation}

These three different strategies have the same theoretical derivatives from semantically equivalent encoder states, but allow different freedoms for optimization.

\section{Experiments}
\label{experiments}
\input{experiments.tex}

\section{Related Work}
Zero-shot translation is of considerable concern among the multilingual translation community. By sharing network parameters across languages, ZS was proven feasible for universal multilingual MT~\cite{Ha2016,Johnson2016}. There are many variations of multilingual models geared towards zero-shot translation. 
\citet{lu2018neural} proposed to explicitly define a recurrent layer with a fixed number of states as ``Interlingua'' which resembles our attention-pooling models. However, they compromise the model compactness by having separate encoder-decoder per language, which linearly increases the model size across languages. On the other hand,  \citet{platanios2018contextual} shares all parameters, but utilized a parameter generator to generate specific parameters for the LSTMs in each language pair using language embeddings.
The closest to our work is probably~\citet{arivazhagan2019missing}. The authors aimed to regularize the model into a common encoding space by taking the mean-pooling of the encoder states and minimize the cosine similarity between the source and the target sentence encodings. In comparison, our approach is more generalized because the decoder is also taken into account during regularization, which is shown by our results on the IWSLT benchmark\footnote{They used the same STAR setup but only reported the average BLEU score across all language pairs}. Also, we proposed stronger representation-forcing since the cosine similarity minimizes the angle between two representational vectors, while the MSE forces them to be exactly equal. 
In addition, zero-resource techniques which generate artificial data for the missing directions have been proposed as an alternative to zero-shot translation~\cite{chen2018zero,al2019consistency,chen2017teacher}. The main disadvantage, however, is the requirement of computationally expensive sampling during training which makes the algorithm less scalable to the number of languages. In our work, we focus on minimally affecting the training paradigm of universal multilingual NMT. 

\section{Conclusion}
This work provides a through investigation of zero-shot translation in multilingual NMT. We conduct an analysis of neural architectures for zero-shot through two three different modifications showing that a beneficial shared representation can be learned for zero-shot translation. Furthermore, we provide a regularization scheme to encourage the model to capture language-independent features for the Transformer model which increases zero-shot performance by $2.23$ BLEU points, achieving the state-of-the-art zero-shot performance in the standard benchmark IWSLT2017 dataset. We also proposed an alternative setting with more than one language as a bridge. 
In this challenging setup for zero-shot translation, we confirmed the consistent effects of our method by showing that the benefit is still significant when languages are far from each other in the pivot path. This result also motivates future works to apply the same strategy for other end-to-end tasks such as speech translation where there may be more variability in domains and modalities.

\section*{Acknowledgments}
\vspace{-0.5em}
The project ELITR leading to this publication has received funding from the European Union’s Horizon 2020 Research and Innovation Programme under grant agreement N\textsuperscript{\underline{o}} 825460.
We thank Elizabeth Salesky for the constructive comments.

\bibliography{naaclhlt2019}
\bibliographystyle{acl_natbib}

\end{document}

%% file: experiments.tex
\subsection{Experimental Setup}
Our experiments use the standard IWSLT2017 benchmark in multilingual translation~\cite{cettolo2017overview}, which established a standardized multilingual corpora in different languages \{English, German, Dutch, Romanian and Italian\}. The data is around $60\%$ true parallel, i.e. the same sentences translated in multiple languages~\cite{dabre2017kyoto}. With the target of zero-shot translation in mind, we designed two different setups which challenge multilingual models but are also industrially practical. 

First, it is typical that English is the most commonly spoken language in the language set, leading the multilingual model to use English as the bridge language participating in all language pairs. Our first setup therefore consists of English$\longleftrightarrow$\{German, Dutch, Italian, Romanian\} language pairs, with $8$ language pairs in total having supervision during training and the remaining $12$ dedicated to the ZS setup.\footnote{This is different from the ZS setup of the IWSLT evaluation campaign in which only 4 out of 20 directions were not present during training.} 

\begin{figure}[htb]
\vspace{-1em}
\centering
\includegraphics[width = 0.5\textwidth]{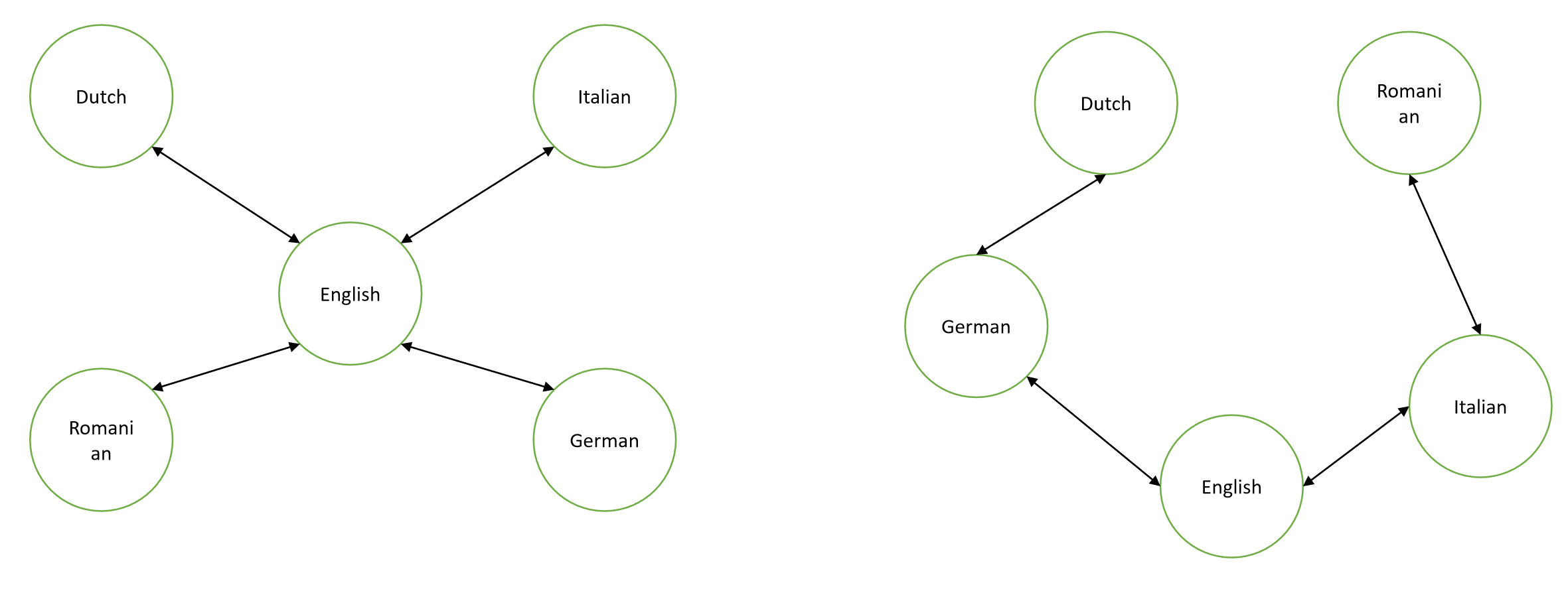}
\caption{\label{fig:setup} The STAR setup (left) with English as the sole bridge language, and the CHAIN setup (right) with 3 different bridge languages and more than 2 steps for zero-shot translation.}
\end{figure}

It is notable that zero-shot (or zero-resource, if the method used generates artificial data to fill the language gap) setups which have been carried out in previous works were mostly concerned language connection with only one bridge (English). However, more realistically, data between local languages or dialects (such as Indian or Vietnamese languages) may more abundant than English. The connectivity in this case demands more than one language for bridging, which is simulated in our second setup by setting a ``Chain'' of languages. This setup also contains 8 supervised languages and $12$ for zero-shot. Figure~\ref{fig:setup} shows the connections between languages in our setups.

The data is preprocessed using standard MT procedures including tokenization and true-casing\footnote{From the Moses toolkit: https://github.com/moses-smt/mosesdecoder} and byte-pair encoding with $40K$ codes. For model selection, the checkpoints performing best on the validation data (dev2010 and tst2010 combined) are averaged, which is then used to translate the tst2017 test set (including all 20 language pairs). 

\subsection{Model Configuration}
Our baseline model is the Transformer following the \textit{Base} configuration in~\cite{vaswani2017attention}. Empirically, we increased the number of layers to $8$ for both encoder and decoder, keeping the layer sizes at $512$ for embedding and $2048$ for the inner layers, and combined with word-dropout of $P_drop=0.1$~\cite{gal2016theoretically} to improve the potency of the baseline for this task. Layer dropout is also added according to the original work with $P_drop=0.2$. The learning rate follows the adaptive learning rate proposed with the Transformer; we use the base learning rate $2$ and the number of warm-up steps is $8192$. The Transformer is trained for around $60000$ steps before overfitting. 

For multilingual functionality, the model uses language embeddings as a feature\footnote{We tried the simpler method with the input token as in \cite{Johnson2016}, but our model could not consistently produce the correct output language in zero-shot tests, which is in-line with~\cite{ha2017effective}}~\cite{ha2017effective}. Our fixed-size models with pooling use $16$ heads for the multi-head pooling models and $16$ attention heads for the attention-pooling models.

\subsection{Training Details}
For all three variations of the decoder, the most important factor is the coefficient~$\alpha$ of the second loss term (as in Equation~\ref{eq:enc_bridge} which decides the importance of this term during the training process. In the beginning of training, it is more important to focus on the main translation tasks, while the regularization term has more effect when the model is converging. To make training stable, repeatable, and reduce the necessity of hyperparameter tuning, we always take the Transformer baseline as the pretrained model and then continue training with the 2\textsuperscript{nd} loss term with constant~$\alpha$. Based on initial experiments for MSE-Decoder and MSE-Attention, we set~$\alpha=0.2$ while it is set to $0.01$ for KL-Softmax. As a result, all of our variations have the same baseline as common ground. Further, when models are trained from the baseline checkpoint, we reset the learning rate and learning rate on an adaptive schedule and continue training for around $50000$ steps\footnote{Prolonging training of the baseline is not beneficial because it will begin to overfit.}. 

An important detail during training is that, it is crucial to free the gradient-path in the decoder from the 2\textsuperscript{nd} loss term for all three variations. In other words, the encoder only receives gradients from regularization. While we saw little difference for the development data with or without this constraint, we noticed that zero-shot translation performance can worsen if gradients flow through the decoder normally\footnote{This can be done in PyTorch by creating a second decoder with (frozen) separate parameters from the main model decoder and then synchronizing them after each update.}. 

Our model is implemented in PyTorch~\cite{paszke2017automatic} and is publicly available~\footnote{The implementation is available at \textit{https://github.com/quanpn90/NMTGMinor/tree/DbMajor}}.

\subsection{Baseline and Fixed-size Source Language Representation Results }
First, as outlined in Section~\ref{sec:pooling}, our goals are to set a competitive baseline and more importantly verify the behaviour of the encoder when the output space is limited to a fixed size instead of variable states. As shown in Table~\ref{tab:pooling}, while the two pooling models suffered from information bottleneck and lost $1-2$ BLEU for each language pair compared to the base Transformer model, the Mean-Pooling model is surprisingly better than the baseline at zero-shot tests. The Attention-Pooling model outperformed the Mean-Pooling at non-zero-shot tests, yet is worse at zero-shot conditions. Compared to other published works on this dataset (which are trained on all 20 directions), our supervised directions set the state-of-the-art for these directions while the zero-shot results approach the best supervised models in the literature~\cite{dabre2017kyoto,platanios2018contextual}). 
Furthermore, by training these two models with a loss function including MSE-loss for encoder similarity, we found noticeable gains on zero-shot performance. More importantly, the zero-shot performance of our Mean-Pooling model with MSE-encoder not only outperforms the baseline, but also rivals the Transformer model competitively trained with all language pairs in~\citet{dabre2017kyoto}. This gain is also noticed with the Attention-Pooling model.

These preliminary experiments shed light on several findings. First, when limited in representation size, the multilingual NMT models can selectively focus on features shared between languages; this is our hypothesis for the improvement in zero-shot translation from the baseline to the \textit{Mean-Pooling} model (on average $+1.28$ BLEU points). Second, by applying the MSE-loss to both pooling variations, they significantly improve in zero-shot performance. This is, to the best of our knowledge, the first empirical proof that a multilingual model is able to learn a common representation space. 

\begin{table*}[htb!]
  \centering
  \begin{tabular}{|c|c|c|c|c|c|c|}
  \hline
  Pair/Model   & Transformer & Mean-Pooling & + MSE & Attn-Pooling & + MSE & Transformer  \\
               &  (ours)    &              &        &              &        & \cite{dabre2017kyoto} \\
  \hline
en-de & 27.51   & 23.74	   &	25.51 & 26.04 & 26.2  & 23.25 \\
de-en &	30.73   & 27.53	   &	28.44 & 28.68 & 29.34 & 26.45 \\
en-ro &	27.45   & 23.48	   &	25.08 & 25.37 & 26.03 & 24.66 \\
ro-en &	33.65   & 30.25	   &	30.95 & 32.1  & 32.02 & 29.58 \\
en-it &	31.84   & 27.71	   &	29.11 & 30.14 & 30.08 & 30.79 \\
it-en &	35.84   & 32.50	   &	33.75 & 34.16 & 34.23 & 34.73 \\
en-nl &	32.15   & 28.58	   &	29.86 & 30.9  & 30.68 & 28.80 \\
nl-en &	34.81   & 31.00	   &	32.1  & 32.81 & 33.02 & 30.49 \\
  \hline
de-nl &	19.04   & 19.68		&	20.46 &	    18.36  & 19.41 & 19.64   \\
nl-de &	20.46   & 19.89	    &	21.10 &	    19.48  & 20.44 & 20.27  \\
it-ro &	18.45   & 18.16		&	19.73 &	    17.42  & 18.74 & 20.60  \\
ro-it &	19.84   & 19.70		&	20.96 &	    18.73  & 19.92 & 21.89  \\
de-it &	16.59   & 16.40		&	17.53 &	    15.23  & 16.59 & 17.54  \\
it-de &	17.55   & 16.91		&	18.89 &	    16.89  & 18.36 & 19.10  \\
nl-ro &	16.89   & 16.63		&	17.85 &	    15.77  & 16.94 & 17.65  \\
ro-nl &	18.12   & 18.65		&	19.79 &	    17.41  & 18.8  & 20.24  \\
nl-it &	18.11   & 18.31		&	19.78 &	    17.45  & 18.54 & 19.86  \\
it-nl &	18.71   & 19.31		&	21.08 & 	18.31  & 19.91 & 22.32  \\
de-ro &	15.33   & 15.07		&	16.13 &	    14.56  & 15.31 & 16.27  \\
ro-de &	17.92   & 17.19	 	&	19.02 &	    17.04  & 18.16 & 17.94  \\
    \hline  
Avg.  &	18.08	&  18.0 	&	19.36        & 17.22  & 18.43  &     \\
 &           &  -0.08    &+\textbf{1.28}  & -0.86  & +0.35  &     \\    
    \hline
  \end{tabular}
  \caption{IWSLT 2017 STAR configuration: Baseline vs (Mean/Attention) Pooling. The top section shows $8$ language pairs involved in training, while the bottom section shows the zero-shot results for $12$ language pairs. We also present results for this dataset from previous work for reference.}
  \label{tab:pooling}
\end{table*}

\subsection{Transformer with Language-Independent Regularization}
In Section~\ref{sec:att-bridge} we showed three different strategies to achieve a decoder that is source language-independent, which theoretically may have the same effect to minimize encoded representation differences: directly equalizing the Softmax outputs, the decoder outputs, and the attention output of each layer. It is important to note that no architectural modification was necessary to include these strategies, thus all of the advantages of the Transformer model and the overall number of parameters are maintained. 



\subsubsection{Results for the STAR Configuration}
\begin{table*}[htb!]
  \centering
  \begin{tabular}{|c|c|c|c|c|c|c|c|}
  \hline
  Pair/Model   & Transformer & +Pivot &  +MSE    & +MSE- & +KL        & Mean-     &  Attn-         \\
               &             &        &  -Attn   &  dec  & -Sofmax   & Pooling   &  Pooling         \\
  \hline
en-de &		27.51	&   &		27.44	&	27.21		&	25.52 &		25.64	&	25.51\\
de-en &		30.73	&	&		30.6	&	30.37		&	29.32 &		29.34	&	28.44\\
en-ro&		27.45	&	&		27.32	&	27.1		&	25.40 &		25.84	&	25.08\\
ro-en&		33.65	&	&		33.24	&	33.62		&	31.89 &		32.12	&	30.95\\
en-it&		31.84	&	&		31.61	&	31.84		&	29.55 &		30.03	&	29.11\\
it-en&		35.84	&	&		35.76	&	35.93		&	34.34 &		34.72	&	33.75\\
en-nl&		32.15	&	&		31.85	&	31.38		&	29.78 &		30.46	&	29.86\\
nl-en&		34.81	&	&		34.3	&	34.52		&	32.97 &		33.25	&	32.1\\
\hline
de-nl	&	19.04	&	21.59	&	20.93	&	21.47	&	19.44 &		20.95	&	20.46\\
nl-de	&	20.46	&	22.14	&	21.99	&	21.9	&	19.93 &		21.51	&	21.1\\
it-ro	&	18.45	&	20.68	&	20.25	&	20.56	&	18.01 &		20.23	&	19.73\\
ro-it	&	19.84	&	22.32	&	21.44	&	22.19	&	20.02 &		21.48	&	20.96\\
de-it	&	16.59	&	19.08	&	18.12	&	18.44	&	17.01 &		18.18	&	17.53\\
it-de	&	17.55	&	20.68	&	19.09	&	19.92	&	18.21 &		19.47	&	18.89\\
nl-ro	&	16.89	&	19.25	&	18.41	&	18.8	&	16.97 &		18.12	&	17.85\\
ro-nl	&	18.12	&	21.38	&	20.16	&	20.8	&	19.33 &		20.34	&	19.79\\
nl-it	&	18.11	&	21.7	&	20.04	&	20.91	&	18.93 &		20.15	&	19.78\\
it-nl	&	18.71	&	22.67	&	21.41	&	21.8	&	19.75 &		21.52	&	21.08\\
de-ro	&	15.33	&	17.69	&	16.77	&	17.12	&	15.47 &		16.56	&	16.13\\
ro-de	&	17.92	&	20.84	&	19.89	&	19.84	&	18.35 &		19.31	&	19.02\\

    \hline  
avg    & 18.08	&	20.83	&     19.88   	&	20.31        &  18.45   &  19.36      &  19.81 \\
 &    &   +2.64   &     +1.80     &+\textbf{2.23}  &  +0.37   &  +1.28    & +1.73  \\    
    \hline
  \end{tabular}
  \caption{IWSLT 2017 STAR configuration result. Here we showed the Mean Pooling model that is enhanced with MSE-Encoder, and the Attn-Pooling model with MSE-Decoder.}
  \label{tab:star}
\end{table*}
The results are shown in table~\ref{tab:star} for the \textit{STAR} configuration. 
Because MSE-attention is the closest derivative to having the same encoder representation, we first investigate the effects of this variation. All zero-shot translation pairs receive noticeable improvement, with the average of $1.71$ BLEU points. The most significant gain belongs to It-Nl pair, which achieves a~$2.7$ BLEU gain. More importantly, unlike the pooling models, we did not have a performance compromise for the non-zero tests. Specifically, the results in the 8 supervised language pairs are nearly identical to the baseline (except for the En-Nl direction, which decreases by $0.8$ points). On average, the benefit for the zero-shot tests greatly outweighs any potential compromise.

The MSE-Decoder allows more freedom during optimization compared to the MSE-attention, as it only requires the final state of the decoder which looks at both encoded sentences to be the same. In this case, we found significant improvement for zero-shot translation with $+2.21$ BLEU points on average. The previously most-improved language pair, It-Nl, is further improved by $0.4$ for a total of $3.1$ BLEU. Moreover, we found that this addition is also helpful for the pooling models, as reflected in the final column of the Table~\ref{tab:star}, significantly increasing the averaged BLEU scores from $17.22$ to $19.81$ points.

Finally, we found that regularizing on Softmax level is extremely difficult to optimize, and the resulting model deteriorates in performance for both zero-shot and normal tests. We found that the gradient norm is much bigger than the other two cases, so possibly optimization can be done with appropriate coefficients. However, this model is the most computationally expensive among the three investigated, due to the second Sofmax function required to be computed, making hyperparameter tuning expensive. 

Even with the significant gain from regularizing the encoder representation, there is still a distance ($0.4$ BLEU point on average) between the best zero-shot model and pivot translation. While pivot translation can theoretically suffer from error cascading, we argue that this is a very strong baseline because the language-specific information, which is possibly negated by finding a language-independent encoder, can be transferred during the pivot process. On the other hand, pivot translation is twice as slow because multiple translation phases are required. 

Our results also proved that our approach does not induce bias toward any language pair, as evidenced by the fact that our improvements (or deterioration) is nearly uniform across language pairs. 

\subsection{Results for the CHAIN configuration}
\begin{table*}[htb!]
  \centering
  \begin{tabular}{|c|c|c|c|c|c|c|}
  \hline
  Pair/Model & Distance  &  Transformer &  Pivot & +MSE-Decoder & +MSE-Attn \\           
  \hline
  						
en-ro&	2& 	21.88 & 24.38 &		24.04 &		23.3 \\
ro-en&	2&	29.82 & 29.29 &		30.79 &		30.92\\
de-it&	2&	17.5  & 19.45 &		19.3  &		18.57\\
it-de&	2&	18.22 & 20.97 &		19.84 &		19.02\\
en-nl&	2&	25.98 & 27.07 &		28.22 &		27.38\\
nl-en&	2&	31.24 & 29.22 &		31.65 &		32.08\\
nl-it&	3&	20.51 & 19.12 &		20.94 &		20.64\\
it-nl&	3&	20.87 & 20.39 &		21.47 &		21.17\\
de-ro&	3&	16.55 & 16.61 &		17.06 &		16.71\\
ro-de&	3&	19.35 & 19.15 &		20.18 &		19.65\\
nl-ro&	4&	16.37 & 16.81 &		17.33 &		16.92\\
ro-nl&	4&	17.55 & 18.55 &		19.66 &		18.88\\
\hline  
Avg.  & all &  19.86 &  20.47 &     \textbf{21.06}&      20.58 \\ 
Avg.  & 2   &  24.10 &  25.06 &     \textbf{25.64}&      25.21 \\ 
Avg.  & 3   &  19.32 &  18.81 &     \textbf{19.91}&      19.54 \\ 
Avg.  & 4   &  16.96 &  17.68 &     \textbf{18.50}&      17.90 \\ 
\hline
    
\hline  

\hline
\end{tabular}
\caption{IWSLT 2017 CHAIN configuration results (12 zero-shot directions).}
\label{tab:chain}
\end{table*}

In this particularly challenging setting where we have multiple bridge languages and different zero-shot distances, we experience different behavior from both pivot techniques and our techniques. 

For the closest language pairs (with 2-step distance), the pivot translation method yields better results than both standard and our methods. The exception is the Romanian-English direction, in which case the pivot language is Italian being closer to Romanian than English. 

It is important to note that most works in the literature used English as the common bridge language; these results indicate that zero-shot performance can be more favourable when language similarity is taken into account. 

When the distance increases, zero-shot translation with forced language-independence using an additional loss clearly outperforms pivot-based translation. We see improvements of more than $1$ BLEU over pivoting for languages with several bridge languages. In this case, both of our techniques still bring improvements for every direction compared to the baseline zero-shot, while potential disadvantages of pivoting, namely error propagation, become clearer.
It is important to note that our regularization techniques scale to settings with multiple bridges. We found the performance enhancement to be most significant for the language pairs which are furthest in the chain (4), with $+1.54$ BLEU points difference compared to the baseline. On the other hand, the Nl$\Longleftrightarrow$It language pairs were most difficult to improve. This is also the setting in which pivot suffered the heaviest loss. To summarize, these multi-steps experiments showed the drawbacks of pivot while at the same time confirm the consistency of our approach. 

%% file: naaclhlt2019.bbl
\begin{thebibliography}{28}
\expandafter\ifx\csname natexlab\endcsname\relax\def\natexlab#1{#1}\fi

\bibitem[{Al-Shedivat and Parikh(2019)}]{al2019consistency}
Maruan Al-Shedivat and Ankur~P Parikh. 2019.
\newblock Consistency by agreement in zero-shot neural machine translation.
\newblock \emph{arXiv preprint arXiv:1904.02338}.

\bibitem[{Arivazhagan et~al.(2019)Arivazhagan, Bapna, Firat, Aharoni, Johnson,
  and Macherey}]{arivazhagan2019missing}
Naveen Arivazhagan, Ankur Bapna, Orhan Firat, Roee Aharoni, Melvin Johnson, and
  Wolfgang Macherey. 2019.
\newblock The missing ingredient in zero-shot neural machine translation.
\newblock \emph{arXiv preprint arXiv:1903.07091}.

\bibitem[{Bahdanau et~al.(2014)Bahdanau, Cho, and Bengio}]{Bahdanau2014}
D.~Bahdanau, K.~Cho, and Y.~Bengio. 2014.
\newblock Neural machine translation by jointly learning to align and
  translate.
\newblock \emph{CoRR}, abs/1409.0473.

\bibitem[{Cettolo et~al.(2017)Cettolo, Federico, Bentivogli, Jan, Sebastian,
  Katsuitho, Koichiro, and Christian}]{cettolo2017overview}
Mauro Cettolo, Marcello Federico, Luisa Bentivogli, Niehues Jan, St{\"u}ker
  Sebastian, Sudoh Katsuitho, Yoshino Koichiro, and Federmann Christian. 2017.
\newblock Overview of the iwslt 2017 evaluation campaign.
\newblock In \emph{International Workshop on Spoken Language Translation},
  pages 2--14.

\bibitem[{Chen et~al.(2017)Chen, Liu, Cheng, and Li}]{chen2017teacher}
Yun Chen, Yang Liu, Yong Cheng, and Victor~OK Li. 2017.
\newblock A teacher-student framework for zero-resource neural machine
  translation.
\newblock \emph{arXiv preprint arXiv:1705.00753}.

\bibitem[{Chen et~al.(2018)Chen, Liu, and Li}]{chen2018zero}
Yun Chen, Yang Liu, and Victor~OK Li. 2018.
\newblock Zero-resource neural machine translation with multi-agent
  communication game.
\newblock In \emph{Thirty-Second AAAI Conference on Artificial Intelligence}.

\bibitem[{Dabre et~al.(2017)Dabre, Cromieres, and Kurohashi}]{dabre2017kyoto}
Raj Dabre, Fabien Cromieres, and Sadao Kurohashi. 2017.
\newblock Kyoto university mt system description for iwslt 2017.
\newblock \emph{Proc. of IWSLT, Tokyo, Japan}.

\bibitem[{Domhan and Hieber(2017)}]{domhan2017using}
Tobias Domhan and Felix Hieber. 2017.
\newblock Using target-side monolingual data for neural machine translation
  through multi-task learning.
\newblock In \emph{Proceedings of the 2017 Conference on Empirical Methods in
  Natural Language Processing}, pages 1500--1505.

\bibitem[{Firat et~al.(2016)Firat, Cho, and Bengio}]{firat2016multi}
Orhan Firat, Kyunghyun Cho, and Yoshua Bengio. 2016.
\newblock Multi-way, multilingual neural machine translation with a shared
  attention mechanism.
\newblock \emph{arXiv preprint arXiv:1601.01073}.

\bibitem[{Gal and Ghahramani(2016)}]{gal2016theoretically}
Yarin Gal and Zoubin Ghahramani. 2016.
\newblock A theoretically grounded application of dropout in recurrent neural
  networks.
\newblock In \emph{Advances in neural information processing systems}, pages
  1019--1027.

\bibitem[{Gu et~al.(2018)Gu, Hassan, Devlin, and Li}]{gu2018universal}
Jiatao Gu, Hany Hassan, Jacob Devlin, and Victor~OK Li. 2018.
\newblock Universal neural machine translation for extremely low resource
  languages.
\newblock \emph{arXiv preprint arXiv:1802.05368}.

\bibitem[{Ha et~al.(2016)Ha, Niehues, and Waibel}]{Ha2016}
Thanh-Le Ha, Jan Niehues, and Alexander Waibel. 2016.
\newblock Toward multilingual neural machine translation with universal encoder
  and decoder.
\newblock In \emph{Proceedings of the 13th International Workshop on Spoken
  Language Translation (IWSLT 2016)}, Seattle, USA.

\bibitem[{Ha et~al.(2017)Ha, Niehues, and Waibel}]{ha2017effective}
Thanh-Le Ha, Jan Niehues, and Alexander Waibel. 2017.
\newblock Effective strategies in zero-shot neural machine translation.
\newblock \emph{arXiv preprint arXiv:1711.07893}.

\bibitem[{He et~al.(2016)He, Xia, Qin, Wang, Yu, Liu, and Ma}]{he2016dual}
Di~He, Yingce Xia, Tao Qin, Liwei Wang, Nenghai Yu, Tie-Yan Liu, and Wei-Ying
  Ma. 2016.
\newblock Dual learning for machine translation.
\newblock In \emph{Advances in Neural Information Processing Systems}, pages
  820--828.

\bibitem[{Johnson et~al.(2016)Johnson, Schuster, Le, Krikun, Wu, Chen, Thorat,
  Viegas, Wattenberg, Corrado, Hughes, and Dean}]{Johnson2016}
M.~Johnson, M.~Schuster, Q.~V. Le, M.~Krikun, Y.~Wu, Z.~Chen, N.~Thorat, F.~B.
  Viegas, M.~Wattenberg, G.~Corrado, M.~Hughes, and J.~Dean. 2016.
\newblock Google{\textquoteright}s multilingual neural machine translation
  system: Enabling zero-shot translation.
\newblock \emph{CoRR}, abs/1611.04558.

\bibitem[{Kalchbrenner and Blunsom(2013)}]{kalchbrenner2013recurrent}
Nal Kalchbrenner and Phil Blunsom. 2013.
\newblock Recurrent continuous translation models.
\newblock In \emph{EMNLP}, volume~3, page 413.

\bibitem[{Kalchbrenner et~al.(2016)Kalchbrenner, Espeholt, Simonyan, Oord,
  Graves, and Kavukcuoglu}]{kalchbrenner2016neural}
Nal Kalchbrenner, Lasse Espeholt, Karen Simonyan, Aaron van~den Oord, Alex
  Graves, and Koray Kavukcuoglu. 2016.
\newblock Neural machine translation in linear time.
\newblock \emph{arXiv preprint arXiv:1610.10099}.

\bibitem[{Kim et~al.(2019)Kim, Gao, and Ney}]{kim2019transfer}
Yunsu Kim, Yingbo Gao, and Hermann Ney. 2019.
\newblock Effective cross-lingual transfer of neural machine translation models
  without shared vocabularies.
\newblock In \emph{Proceedings of the Annual Meeting on Association for
  Computational Linguistics (ACL 2019)}.

\bibitem[{Lu et~al.(2018)Lu, Keung, Ladhak, Bhardwaj, Zhang, and
  Sun}]{lu2018neural}
Yichao Lu, Phillip Keung, Faisal Ladhak, Vikas Bhardwaj, Shaonan Zhang, and
  Jason Sun. 2018.
\newblock A neural interlingua for multilingual machine translation.
\newblock \emph{arXiv preprint arXiv:1804.08198}.

\bibitem[{Luong et~al.(2015)Luong, Pham, and Manning}]{luong2015effective}
Minh-Thang Luong, Hieu Pham, and Christopher~D Manning. 2015.
\newblock Effective approaches to attention-based neural machine translation.
\newblock \emph{arXiv preprint arXiv:1508.04025}.

\bibitem[{Niehues and Cho(2017)}]{niehues2017exploiting}
Jan Niehues and Eunah Cho. 2017.
\newblock Exploiting linguistic resources for neural machine translation using
  multi-task learning.
\newblock In \emph{Proceedings of the Second Conference on Machine
  Translation}, pages 80--89.

\bibitem[{Paszke et~al.(2017)Paszke, Gross, Chintala, Chanan, Yang, DeVito,
  Lin, Desmaison, Antiga, and Lerer}]{paszke2017automatic}
Adam Paszke, Sam Gross, Soumith Chintala, Gregory Chanan, Edward Yang, Zachary
  DeVito, Zeming Lin, Alban Desmaison, Luca Antiga, and Adam Lerer. 2017.
\newblock Automatic differentiation in pytorch.

\bibitem[{Platanios et~al.(2018)Platanios, Sachan, Neubig, and
  Mitchell}]{platanios2018contextual}
Emmanouil~Antonios Platanios, Mrinmaya Sachan, Graham Neubig, and Tom Mitchell.
  2018.
\newblock Contextual parameter generation for universal neural machine
  translation.
\newblock \emph{arXiv preprint arXiv:1808.08493}.

\bibitem[{Schwenk and Douze(2017)}]{schwenk2017learning}
Holger Schwenk and Matthijs Douze. 2017.
\newblock Learning joint multilingual sentence representations with neural
  machine translation.
\newblock \emph{arXiv preprint arXiv:1704.04154}.

\bibitem[{Sutskever et~al.(2014)Sutskever, Vinyals, and Le}]{Sutskever2014}
I.~Sutskever, O.~Vinyals, and Q.~V. Le. 2014.
\newblock Sequence to sequence learning with neural networks.
\newblock In \emph{Advances in Neural Information Processing Systems 27: Annual
  Conference on Neural Information Processing Systems 2014}, pages 3104--3112,
  Quebec, Canada.

\bibitem[{Vaswani et~al.(2017)Vaswani, Shazeer, Parmar, Uszkoreit, Jones,
  Gomez, Kaiser, and Polosukhin}]{vaswani2017attention}
Ashish Vaswani, Noam Shazeer, Niki Parmar, Jakob Uszkoreit, Llion Jones,
  Aidan~N Gomez, Lukasz Kaiser, and Illia Polosukhin. 2017.
\newblock Attention is all you need.
\newblock \emph{arXiv preprint arXiv:1706.03762}.

\bibitem[{Wang et~al.(2017)Wang, Finch, Utiyama, and
  Sumita}]{wang-etal-2017-sentence}
Rui Wang, Andrew Finch, Masao Utiyama, and Eiichiro Sumita. 2017.
\newblock \href {https://doi.org/10.18653/v1/P17-2089} {Sentence embedding for
  neural machine translation domain adaptation}.
\newblock In \emph{Proceedings of the 55th Annual Meeting of the Association
  for Computational Linguistics (Volume 2: Short Papers)}, pages 560--566,
  Vancouver, Canada. Association for Computational Linguistics.

\bibitem[{Zoph et~al.(2016)Zoph, Yuret, May, and Knight}]{zoph2016transfer}
Barret Zoph, Deniz Yuret, Jonathan May, and Kevin Knight. 2016.
\newblock Transfer learning for low-resource neural machine translation.
\newblock In \emph{Proceedings of the Conference on Empirical Methods in
  Natural Language Processing (EMNLP 2016)}, pages 1568--1575, Austin, USA.

\end{thebibliography}
